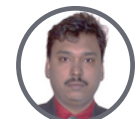

Rajesh Malik

# Energy saving in off-road vehicles using leakage compensation technique

Gyan Wrat*, J. Das**

Aalborg University, Denmark *
IIT(ISM), Dhanbad, India **

The article focuses on enhancing the energy efficiency of linear actuators used in heavy earth moving equipment, particularly in the booms of excavation equipment. Two hydraulic circuits are compared in terms of energy efficiency, with one using a conventional proportional directional control valve (PDCV) and the other using an innovative solution of proportional flow control valve (PFCV) with artificial leakage between the two ends of the actuator. The PFCV reduces energy loss in the form of heat by bypassing the extra flow from the pump during position control, unlike the PDCV that uses a pressure relief valve. The hydraulic circuit using PFCV is found to be 8.5% more energy efficient than the conventional circuit using PDCV.

The article also discusses the position control of the actuator, which is achieved using a PID controller tuned by a fuzzy controller. The simulation of the hydraulic circuit is carried out using MATLAB/Simulink, and the results are compared with experiments. Overall, the proposed approach could lead to significant improvements in the energy efficiency of linear actuators used in heavy earth moving equipment, thereby reducing their environmental impact and operating costs.

## 1. Introduction

This This article discusses the issue of low efficiency in hydraulic systems used in construction and mining industries. The constant hydraulic power supply in conventional systems leads to inefficiencies, and additional losses caused by throttling and overflowing further reduce efficiency [1-5]. Energy-saving and emission-reducing designs are therefore crucial in these hydraulic systems [6-9]. The article focuses on developing an energy-efficient position control strategy for linear actuators used in construction equipment.

The authors review recent literature on energy-efficient hydraulic circuits used in heavy earth-moving equipment. Tianliang et al. [9] propose a new type of proportional relief valve (PRV) connected to a hydraulic energy regeneration unit (HERU) to overcome traditional relief valve energy losses. Their experimental results show improved performance and energy savings compared to traditional PRVs. Ali et al. [10] address the control of actuator position and supply pressure simultaneously in an electro-hydraulic servo system (EHSS), reducing overall energy consumption.

Other researchers explore energy-saving techniques using load sensing pumps. Chiang and Chien [11] compare three different systems and find that load sensing systems save about 61.2% energy compared to conventional systems. Wang and Wang [12] focus on minimizing overflow and excess pressure losses using variable supply pressure (VSP) control. Baghestan et al.[13] propose an energy-saving position control strategy for EHSS, employing different control laws for proportional directional valve and pressure relief valve.

In their research, Wrat et al. [14] investigated the impacts of two control strategies on position control of an actuator. These strategies involved controlling the electric motor speed and the swash plate angle of the main pump. Through simulations and tests, it was observed that the swash plate control strategy demonstrated superior response and dynamic characteristics compared to alternative strategies. However, it is important to note that this article did not address the aspect of energy conservation. Xu and Cheng [15] surveyed the impact of energy-saving technologies and control solutions in multi-actuator hydraulic systems. It was discovered that employing advanced coordinated position control

(CPC) or coordinated rate control (CRC) can attain comfort, safety, precision, and efficient operation. In the context of mobile hydraulics, motion control is utilized to achieve desirable characteristics such as reduced environmental impact, enhanced energy efficiency, effective operation, and optimal dynamic performance.

The article also covers control strategies for actuator position using electric motor speed and swash plate angle control, highlighting the importance of coordinated position control (CPC) and coordinated rate control (CRC). Some researchers investigate meter-in and meter-out flow control methods for energy savings. De Boer and Yao [16] study velocity control using a programmable valve, while Ding et al. propose a combined pressure/flow hybrid pump control with meter-out (MO) valve control, achieving significant energy savings.

Research has also been conducted on hybrid hydraulic transmissions and regeneration devices. Bhola et al. [17] develop a hybrid hydraulic transmission for a front-end loader using an electric generator and battery bank. Triet and Ahn [18] propose a hydrostatic energy-efficient system using a flywheel as an energy capturing device. Wang and Wang [19]design an energy-efficient system for cranes, incorporating a pressure-compensated hydraulic circuit and electrical regeneration device.

Other energy-saving approaches include a novel electro-hydraulic energy-saving system, a hydraulic hybrid luffing system, and leaking valve-pump parallel control. Fuzzy-PID controllers are used to improve dynamic response, and the article compares them to conventional PID controllers.

The article concludes by addressing the lack of research on flow control valves in the context of energy savings. A comparison between two-position control strategies using a proportional directional control valve (PDCV) and a flow control valve (FCV) is conducted, revealing that the FCV is 8.5% more efficient. The stability analysis of the actuator is suggested as a future scope of work.

The remaining sections of the article provide detailed descriptions of the hydraulic system, governing equations, experimental setup,

# hyloc hydrotechnic pvt. ltd.

## Tube Couplings

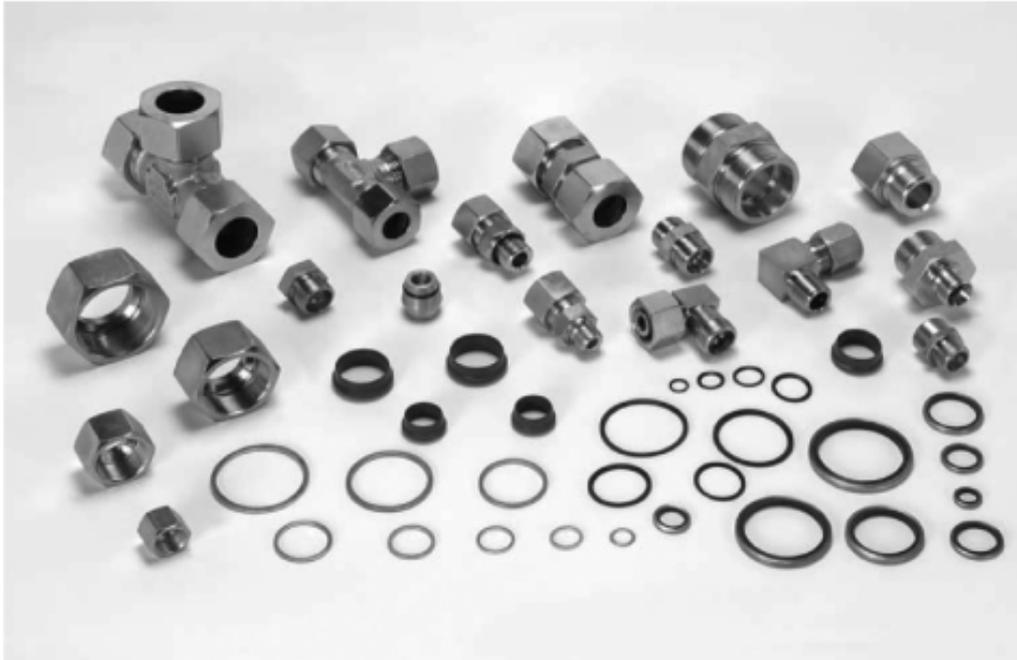

Hyloc is a trusted brand in high quality tube couplings as per ISO 8434/DIN 2353 in steel and stainless steel. Sizes available are 6mm to 42mm in Light and 6mm to 38mm in Heavy series.

## Hydraulic Valves / Flanges / Tube and Pipe Clamps and Diagnostic couplings

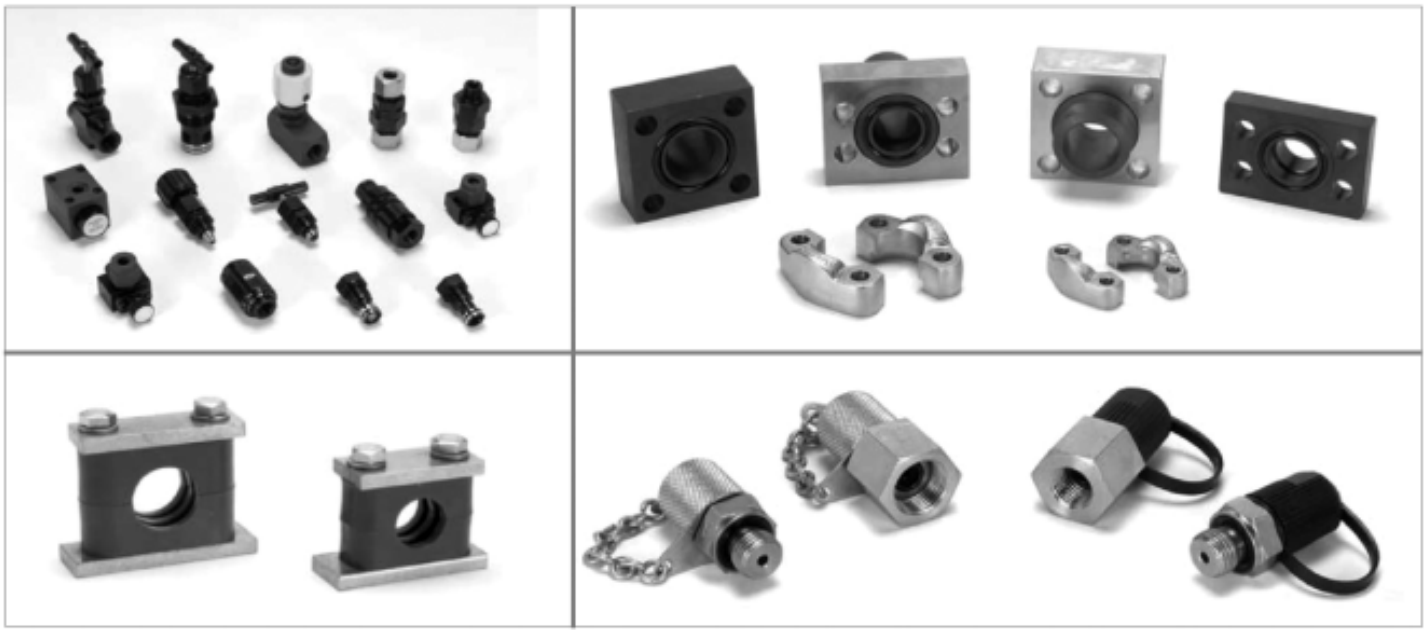

For more details about our products, please contact our Marketing team

marketing@hyloc.co.in

## hyloc hydrotechnic pvt. ltd.

ISO 9001, ISO 14001and OHSAS 18001 Company
DNV Certified TYPE APPROVED SUPPLIER for Marine

88, Machhe Industrial Estate, Machhe, Belgaum 590 014 - INDIA
Tel : +91 831 241 2777          Fax : +91 831 241 2784
E-mail : sales@hyloc.co.in       Web : www.hyloc.co.in

system simulation modeling assumptions, fuzzy-PID controller structure, and important results obtained from two test cases. Finally, the article concludes with a summary of the findings.

## 2. System description

The article presents a test set-up for analyzing the circuit diagram, as shown in Figure 1. The set-up includes an induction motor (1) driving the main pump (2), with the flow-meter (10) measuring the flow-rate to the main cylinder (8). The position control of the cylinder is achieved using the 4/3 PDCV (4), while the FCV (9) is kept off and the on/off DCV (15) is on during the first mode. In the second mode, the PDCV (4) is fully opened, and the FCV (14) controls the flow through the cylinder for position control.

The throttle valve is used for creating back pressure during retraction of the cylinder, controlled by the on/off DCV (15). Pressure transducers (5.1) and (5.2) measure pressures at the rod and bore end sides, as well as the main pump outlet pressure. The main pressure relief valve (PRV) (11) bypasses excess flow during position control.

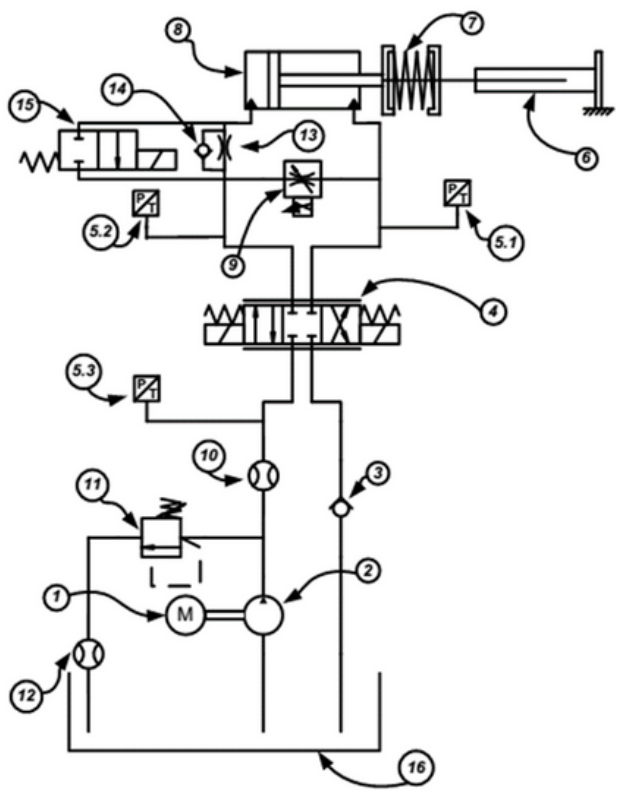

**Figure 1:** Hydraulic circuit diagram for the test set-up (source reference [20])

## 3. Mathematical Modelling

Below, the mathematical models for the two hydraulic systems are presented, starting from equation (1) to equation (11). The system equations are derived from MATLAB/Simulink models, representing two different configurations of hydraulic circuits: one utilizing the Proportional Direction Control Valve (PDCV) and the other utilizing the Flow Control Valve (FCV).

$$Q_1 = Av + \frac{P_1 - P_2}{R_{lkga}} + \frac{V}{\beta}\frac{dP_1}{dt} \qquad (1)$$

$$Av + \frac{P_1 - P_2}{R_{lkga}} - \frac{V}{\beta}\frac{dP_2}{dt} = Q_2 \qquad (2)$$

$$P_L = P_1 - P_2 \qquad (3)$$

$$Q_L = \frac{Q_1 + Q_2}{2} \qquad (4)$$

$$Q_L = \frac{Q_1 + Q_2}{2} \qquad (4)$$

$$Q_L = Av + \frac{P_L}{R_{lkga}} + \frac{V}{2\beta}\frac{dP_L}{dt} \qquad (5)$$

$$F = (P_1 - P_2)A = M\frac{dv}{dt} + \beta_v v + KX \qquad (6)$$

$$Q_L = K_q x_v - K_c P_L \qquad (7)$$

1. Equation (1) describes the flow (Q1) in the actuator, considering actuator movement, leakage through cylinder clearances, and compressible flow in the cylinder.
2. Equation (2) represents the flow (Q2) coming out of the cylinder, balancing actuator motion, leakage, and compressible fluid in the cylinder.
3. Equation (3) calculates the load pressure (PL) of the actuator, which is the difference between the inlet pressure (P1) and outlet pressure (P2).
4. Equation (4) determines the load flow (QL) inside the actuator, which is the average of the inlet flow (Q1) and outlet flow (Q2).
5. Equation (5) provides the load flow equation (QL) inside the actuator, derived from the previous equations.
6. Equation (6) represents the forces acting on the cylinder, including pressure force, inertial force, viscous friction, and external load.
7. Equation (7) relates the load flow (QL) to the spool position and load pressure (PL).

For the hydraulic system using FCV (source reference [20]):

$$Q_0 = C_d A \sqrt{\frac{2P_1}{\rho}} = K_q x_v - K_L P_1 \qquad (8)$$

$$Q_1 = Q_0 + Q_2 \qquad (9)$$

$$Q_1 = K_q x_c - K_c P_1 + Av + \frac{V}{\beta}\frac{dP_1}{dt} + \frac{P_1}{R_{lkga}} \qquad (10)$$

$$P_1 A = M\frac{dv}{dt} + \beta_v v + KX \qquad (11)$$

1. Equation (8) calculates the flow (Q0) through the Flow Control Valve (FCV), based on the spool position and inlet pressure (P1) of the cylinder.
2. Equation (9) expresses the total flow (Q1) from the pump, which is the sum of the flow through the FCV (Q0) and the flow through the cylinder (Q2).
3. Equation (10) determines the flow (Q1) considering the spool position, load pressure (P1), actuator flow (Q1), leakage, and compressible flow.
4. Equation (11) calculates the hydraulic force at the cylinder end, considering mass acceleration, viscous friction, and external load.

The hydraulic system depicted in Figure 2 is based on the use of PFCV (Proportional Flow Control Valve). In this configuration, the PDCV (Proportional Direction Control Valve) is fully open, resulting in atmospheric back pressure on the rod end during extension and the bore end during retraction. To regulate the actuator's position, the excess flow is directed through the PFCV, allowing for precise control.

Figure 5 depicts the schematic representation of the hydraulic system employing PDCV (Pressure-Dependent Control Valve). In this configuration, the flow directed towards the actuator is regulated by adjusting the opening area of the PDCV. Additionally, any excess flow is diverted through the pressure relief valve once it reaches the cracking pressure.

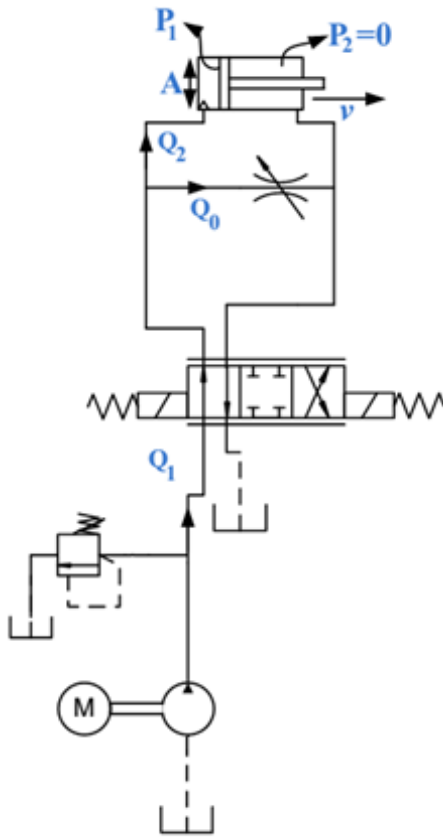

**Figure 2:** The circuit diagram depicts the use of a Flow Control Valve (PFCV) during the extension of an actuator (source reference [20])

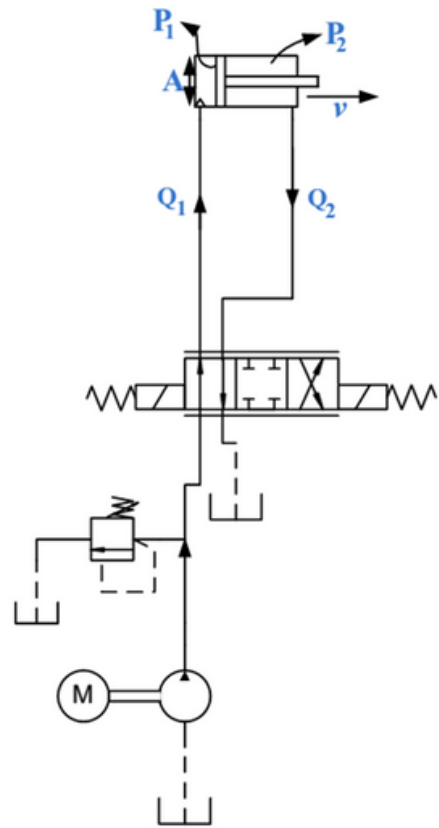

**Figure 5:** The schematic diagram illustrating the use of a Proportional Directional Control Valve (PDCV) during the extension of an actuator (source reference [20])

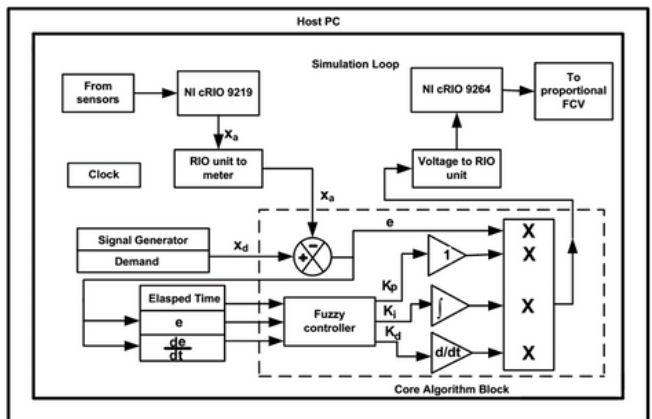

**Figure 3:** The position control strategy implemented in LabVIEW utilizes the Proportional Flow Control Valve (PFCV) (see reference [20])

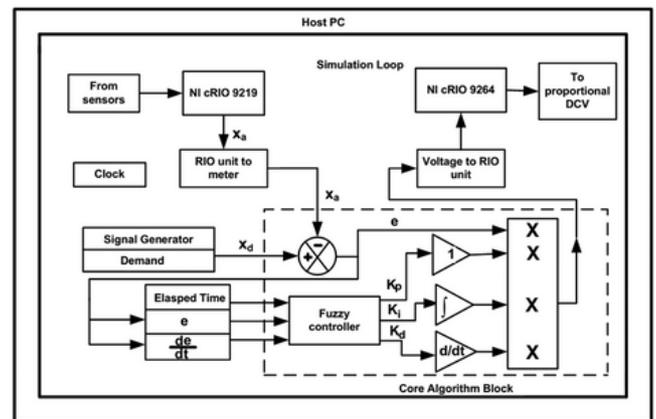

**Figure 6:** Position control strategy in LabVIEW using the Proportional Directional Control Valve PDCV (source reference [20])

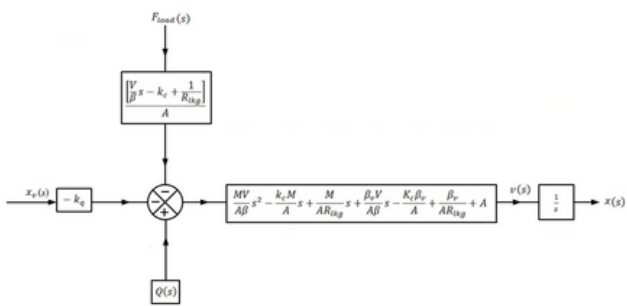

**Figure 4:** Transfer function of the system using PFCV (source reference [20])

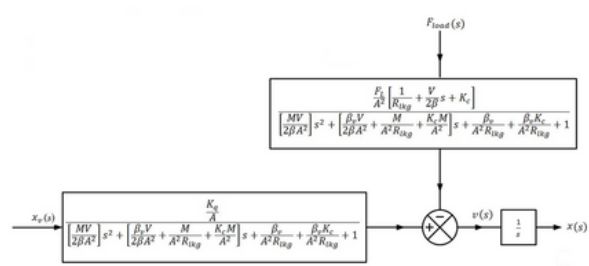

**Figure 7:** The transfer function of the system using (PDCV) (source reference [20])

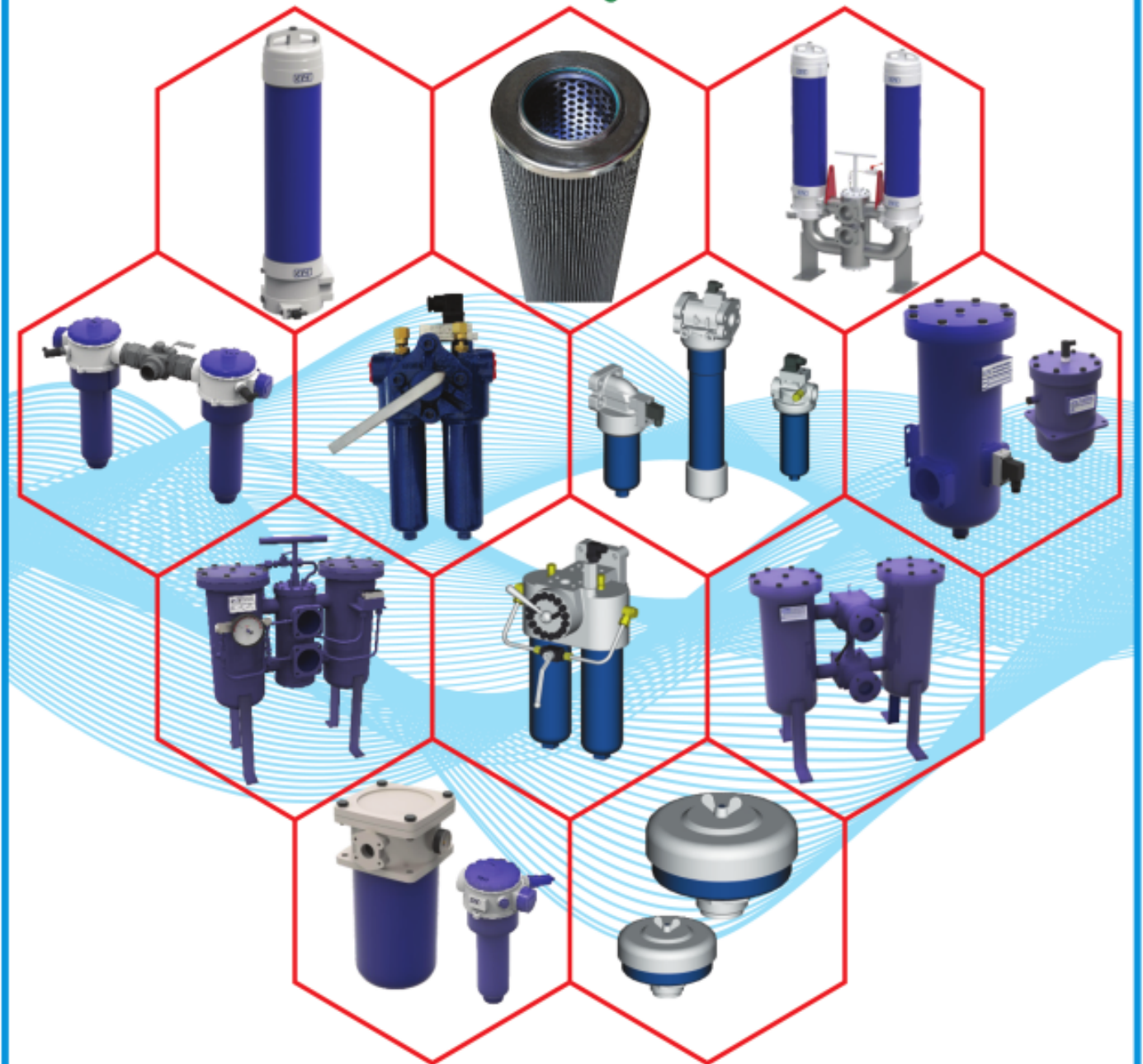

Creating a
*Successful*
Tomorrow

## EPE PROCESS FILTERS & ACCUMULATORS PVT LTD

**Techni Towers**
C-54/A, TSIIC, Balanagar, Hyderabad -500 037. Telangana, India. Tel. Nos. : 0091-40-23778803/23778804/23871445. Fax No. : 0091-40-23871447.
Internet : www.epe-india.com  E-mail  : business@epe-india.com

## 4. Experimental description

The experiments described in figures 3 and 6 were conducted using LabVIEW14 software and the National Instruments controller CompactRIO (NI-cRIO). The LabVIEW14 software, which serves as a Graphic User Interface (GUI) software, was connected to the host PC through the 16-bit NI-cRIO 9076 real-time processor, along with the input module NI-cRIO 9219 and output module NI-cRIO 9264. Within the LabVIEW environment, a Virtual Instrumentation (VI) program was created and linked to the FPGA (Field Programmable Gate Array) module of the controller. The sensors provided real-time analog signals ranging from 0-10V, and the output signals from the analog output module were used to control the PDCV and PFCV for actuator position control, as well as the meter-out orifice for applying load to the actuator by generating back pressure during extension and retraction.

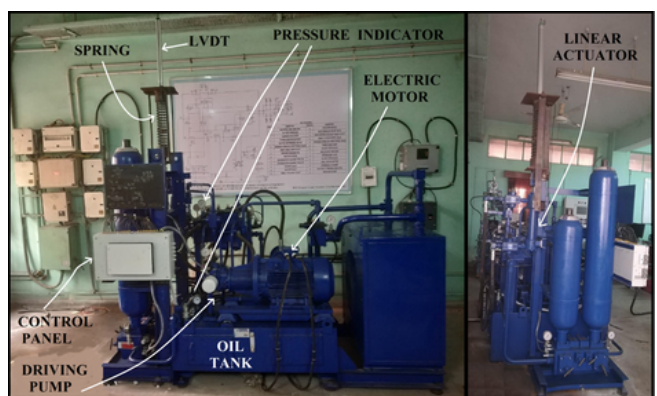

**Figure 8:** Experiment setup (source reference [20])

## 5. System modelling

During the simulation of the hydraulic circuit in MATLAB/Simulink, the following assumptions were made:

- The inertia of the fluid was neglected in this study.
- The effects of temperature and pressure on fluid properties, such as leakage resistances and compressibility, were not considered.
- The valve opening was taken into account, but the effects of system parameters on valve flow and pressure coefficient were disregarded.
- A constant inertial force was assumed in the simulation.

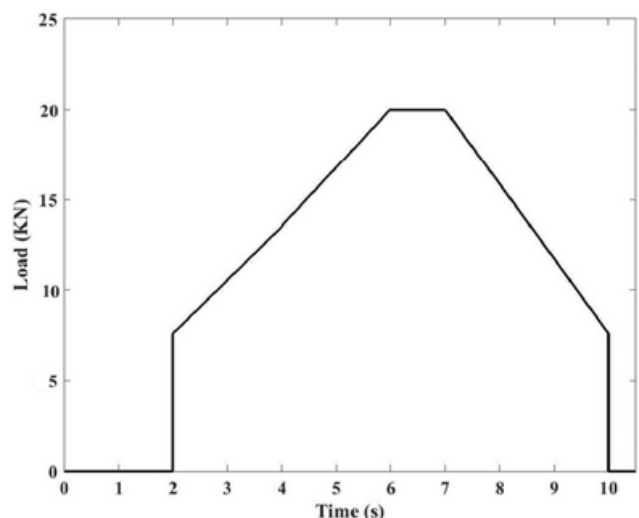

**Figure 9:** The simulated load applied to the cylinder (source reference [20])

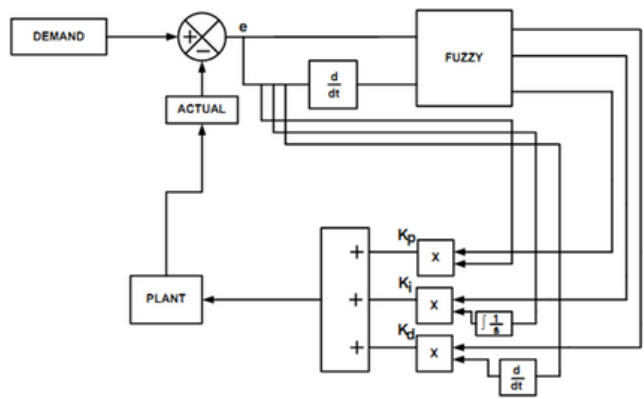

**Figure 10:** Block diagram for closed loop with Fuzzy-PID (source reference [20])

The control strategy utilized for online tuning of the three control parameters Kp, Ki, and Kd of the PID using a fuzzy controller is illustrated in Figure 10. The control scheme employs two inputs: the error and the rate of change of the error.

By utilizing predefined rules and defined input and output ranges, the fuzzy controller dynamically adjusts the parameters online. This adaptive tuning approach effectively mitigates significant nonlinear effects within the system, leading to improved control performance in comparison to conventional PID control methods.

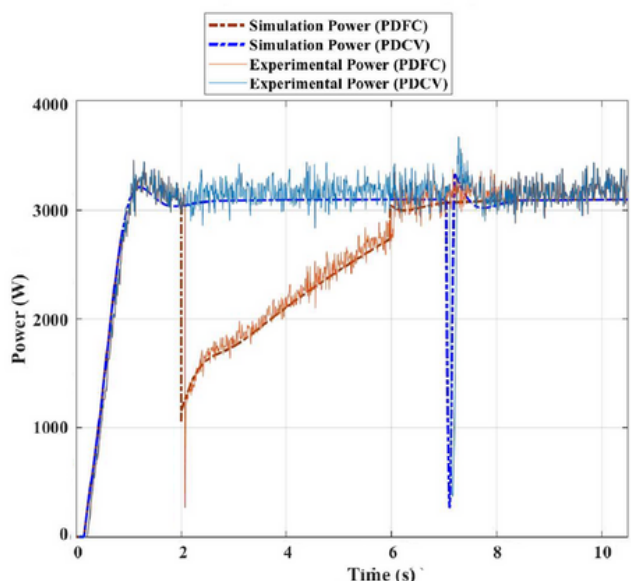

**Figure 11:** Power comparison of two hydraulic circuit (source reference [20])

## 7. Results and discussion
**Energy Consumption comparison**

The energy consumption of both hydraulic systems can be represented by the equation provided (source reference [20])

$$P = \int_{t}^{t+\Delta t} pQ\,dt \qquad (14)$$

where,
P is power outlet by the pump,
p is the measurement of pressure sensor 5.3,
Q is the measurement of flow sensor 10,
t is the starting time of the experiment,
Δt is time duration of the experiment cycle.

| System | Energy Consumption (KJ) |
|---|---|
| Conventional system with PDCV | 30.47 |
| Proposed system with PFCV | 27.867 |
| Energy Saving | 8.54% |

**Table 1:** Energy Consumption comparisons of two systems (source reference [20])

The energy consumption of two hydraulic circuits is compared in Figure 11. The circuit using PDCV exhibits higher energy consumption compared to the one using PFCV. This is primarily attributed to the fact that the excess flow in the PDCV circuit is discharged through the pressure relief valve once it reaches its cracking pressure, as depicted in Figure 15. On the other hand, in the PFCV circuit, the excess flow, as shown in Figure 12, is bypassed at the load pressure, resulting in an efficiency increase of approximately 8.54%. This improvement in efficiency is presented in a tabular format in Table 1.

## 8. Conclusion
This paper conducts a comparative analysis of energy-saving position control methods using two different strategies: proportional directional control valve (PDCV) and flow control valve (FCV) connected between the cylinder ends. The simulation model is created in MATLAB/Simulink, and experimental tests are conducted using LabVIEW software. The experimental results reveal that the FCV strategy, which connects the flow control valve between the cylinder ends, exhibits 8.54% higher efficiency compared to the position control achieved through the proportional directional control valve. The actuator's position control is achieved using a PID-fuzzy control strategy implemented with the proportional directional valve and flow control valve. This control and design approach offers a valuable tool for engineers to develop more energy-efficient and reliable actuators. Further research can focus on analyzing the actuator's response in terms of vibration and developing intelligent control techniques to suppress oscillations during cylinder control.

## Acknowledgement
The authors would like to express their gratitude to the Indian Institute of Technology (Indian School of Mines), Dhanbad for providing the experimental setup. This work has received support from DST (Project Grant Number YSS/2015/000397), and the authors acknowledge this support.

*For the complete list of references, please contact the Editorial staff*